\def\year{2017}\relax
\documentclass[letterpaper]{article}
\usepackage{aaai17}
\usepackage{times}
\usepackage{helvet}
\usepackage{courier}
\usepackage{amsmath}
\usepackage{color}
\usepackage{booktabs}       
\usepackage{amssymb}
\usepackage{graphicx}
\usepackage{multirow}
\usepackage{array}
\usepackage[tight,normalsize,sf,SF]{subfigure}
\usepackage[ruled,vlined]{algorithm2e}
\renewcommand{\KwResult}{\textbf{Output:}}
\renewcommand{\KwData}{\textbf{Input:}}

\def\etc{\emph{etc}.}
\def\eg{\emph{e.g.}}
\def\ie{\emph{i.e.}}
\def\X{\mathcal{X}}
\def\J{\mathcal{J}}
\def\V{\mathcal{V}}
\def\Y{\mathcal{Y}}
\def\Z{\mathcal{Z}}
\def\B{\mathcal{B}}
\def\F{\mathcal{F}}

\begin{document}
%
\title{Multidimensional Scaling on Multiple Input Distance Matrices}
\author{Song Bai$^1$, Xiang Bai$^{1}$\thanks{Corresponding author}, Longin Jan Latecki$^2$, Qi Tian$^3$\\
$^1$Huazhong University of Science and Technology\\
$^2$Temple University,~~$^3$University of Texas at San Antonio\\
\{songbai, xbai\}@hust.edu.cn,~~latecki@temple.edu,~~qitian@cs.utsa.edu}
\maketitle
\begin{abstract}
Multidimensional Scaling (MDS) is a classic technique that seeks vectorial representations for data points, given the pairwise distances between them. In recent years, data are usually collected from diverse sources or have multiple heterogeneous representations. However, how to do multidimensional scaling on multiple input distance matrices is still unsolved to our best knowledge.

In this paper, we first define this new task formally. Then, we propose a new algorithm called Multi-View Multidimensional Scaling (MVMDS) by considering each input distance matrix as one view.
The proposed algorithm can learn the weights of views (\ie,~distance matrices) automatically by exploring the consensus information and complementary nature of views.
Experimental results on synthetic as well as real datasets demonstrate the effectiveness of MVMDS.
We hope that our work encourages a wider consideration in many domains where MDS is needed.
\end{abstract}

\section{Introduction}
Multidimensional scaling (MDS)~\cite{borg2005modern,torgerson1958theory} is a fundamental and important technique with a wide range of applications to data visualization, artificial intelligence, network localization, robotics, cybernetics, social science,~\etc~For example, researchers in bioinformatics apply MDS to unravel relational patterns among genes~\cite{taguchi2005relational}. As another example, MDS is also used by computer vision community~\cite{bronstein2008analysis}. A typical application is to approximate geodesic distances of mesh points~\cite{elad2003bending} or planar points~\cite{IDSC} in Euclidean space so that the non-rigid intrinsic structure of shapes can be captured.

Given pairwise distances between $N$ data points, MDS aims at projecting these data into $P$ dimensional space, such that the between-object distances can be preserved as well as possible.
In recent years, data are often collected from diverse domains or have various heterogeneous representations~\cite{amid2015multiview,xu2013survey}. That is to say, each data may have multiple views.
For instance, an image can be described by multiple visual features, such as Scale Invariant Feature Transform (SIFT)~\cite{SIFT}, Histogram of Oriented Gradients (HOG)~\cite{HOG}, Local Binary Patterns (LBP)~\cite{LBP},~\etc~A web page can be described by the document text itself and the anchor text attached to its hyperlinks. It has been extensively demonstrated that a fusion of those multi-view representations by leveraging the interactions and complementarity between them is usually beneficial to obtain more faithful and accurate information.

In the past decades, numerous efforts have been devoted to the formulation, optimization and application of MDS (see~\cite{borg2005modern,france2011two} for a survey). However, the problem of multidimensional scaling on multiple input distance matrices has not been addressed. Nevertheless, this new topic has gradually become important in practical applications. Consider a toy example (also presented in experiments) where one wants to illustrate the relative positions of six cities in a planar map but he/she receives more than one distance matrix. How to project the six cities to $P=2$ dimensional space given the multiple input matrices? Meanwhile, MDS can also act as a dimensionality reduction algorithm if the embedding dimension $P$ is smaller than the input dimension. This arises another question, that is, how to conduct multi-view dimensionality reduction~\cite{MVDR1,MVDR2,foster2008multi} with the rapid growth of high dimensional data.

In this paper, we begin to investigate this new task,~\ie,~multidimensional scaling on multiple input distance matrices. Our contributions can be divided into three folds:
\begin{enumerate}
  \item We formally put forward the idea of performing MDS on multi-view data, and discuss the basic difficulties that need to be addressed carefully in this framework.
  \item In addition to the novelty of our problem formulation, a new algorithm called Multi-View Multidimensional Scaling (MVMDS) is proposed to solve it. Inspired by~\cite{MVDR2} and its related works in multi-view learning~\cite{sun2013survey}, a weight learning paradigm is imposed to attach more importance to discriminative views and suppress the negative influences of noisy views. Accordingly, an iterative solution is derived with proven convergence so that view weights can be updated automatically controlled with only one parameter.
  \item Extensive experimental evaluations on synthetic and real datasets manifest the effectiveness of the proposed method. Besides, we also give a comprehensive summary about promising future works that can be studied within this framework.
\end{enumerate}

\section{Task Definition}
Given the pairwise distances $\delta=\{\delta_{ij}\}_{1\leq i,j\leq N}$ between $N$ data points, Multidimensional Scaling (MDS) seeks for $N$ configuration points $\X=\{x_1,x_2,\dots,x_N\}\in\mathbb{R}^{N\times P}$ such that $\delta_{ij}$ can be well approximated by the Euclidean distance $d_{ij}(\X)=\|x_i-x_j\|_2$. The most widely-used definition of metric MDS is called ``Stress", defined as
\begin{equation} \label{eq:mds}
\min_{\X}\sum_{i<j}w_{ij}\left(\delta_{ij}-d_{ij}(\X)\right)^2,
\end{equation}
where $w_{ij}$ are some pre-fixed weighting coefficients and $P$ is the embedding dimension. In some specific situations, we have to deal with missing values,~\ie,~$\delta_{ij}$ is not well defined. Therefore, one can set $w_{ij}$ to $0$ for those missing values, and set $w_{ij}$ to $1$ if $\delta_{ij}$ is known.

As discussed above, though numerous efforts have been devoted to its optimization and application, all the variants of MDS can only deal with single view data. Performing MDS in multi-view data has not been addressed. In this paper, we first give a formal definition of performing MDS on multi-view data. Given $N$ abstract points $\Y=\{y_1,y_2,\dots,y_N\}$ and their pairwise distances in $M$ views $\delta^{(v)}\in\mathbb{R}^{N\times N}, 1\leq v\leq M$, where our goal is to learn a function $\F$ defined as
\begin{equation}\
\F:(\Y,\delta^{(1)},\delta^{(2)},\dots,\delta^{(M)})\rightarrow \X, M>1,
\end{equation}
where $\X\in\mathbb{R}^{N\times P}$ is a configuration of $N$ in $P$ dimensional Euclidean space.

In this framework, some basic difficulties should be solved carefully: (1) The most fundamental issue is how to ensemble these distance matrices. A very naive solution is to use a linear combination of them. However, it is likely to achieve unsatisfactory results since informative and noisy views are all treated equally. Another possible solution is to apply co-training~\cite{co_training2} to MDS. Unfortunately, many co-training algorithms cannot guarantee the convergence. (2) How to judge the importance of different views? In most cases, MDS is defined as an unsupervised algorithm. It is problematic to determine the view weights automatically in an unsupervised manner, since no prior knowledge is available. (3) How to derive the optimal solution from $\F$ which guarantees to provide meaningful results?

\section{Proposed Solution} \label{sec:proposed}
To do multidimensional scaling on multiple input distance matrices, we propose a new objective function called Multi-View Multidimensional Scaling (MVMDS), formulated as
\begin{equation} \label{eq:mvmds}
\begin{split}
& \min_{\alpha^{(v)},\X}\sum_{v=1}^M{\alpha^{(v)}}^\gamma\sum_{i<j}w_{ij}\left(\delta_{ij}^{(v)}-d_{ij}(\X)\right)^2, \\
& s.t.~\sum_{v=1}^M{\alpha^{(v)}}=1, 0\leq \alpha^{(v)}\leq 1,
\end{split}
\end{equation}
where $\alpha^{(v)}$ measures the importance of $v$-th view, and the exponent $\gamma>1$ is the weight controller that determines the distribution of $\alpha=\{\alpha^{(1)},\alpha^{(2)},\dots,\alpha^{(M)}\}$.

The weight learning mechanism is imposed by adding ${\alpha^{(v)}}^\gamma$ to the stress. The reason behind this choice is that if using $\alpha^{(v)}$ directly, the solution of $\alpha$ is that the view with the smallest stress value has the weight $\alpha^{(v)}=1$ and all other views have $\alpha^{(v)}=0$. This is not a good behavior since only one view is selected and the complementary nature among multiple views is ignored. The proposed adaptive weight learning paradigm is a primary advantage over the naive solution of using a weighted linear combination of multiple distance matrices, where it is nontrivial to determine the weights since at least $M-1$ values should be specified. Hence the computational complexity is unbearable when $M>2$.

Meanwhile, we only set a consensus embedding $\X$, instead of defining an individual embedding $\X^{(v)}$ for each view. It can be understood as "minimizing disagreement?of multiple views. Nevertheless, we force the embedding $\X$ to be the same across multiple views so that the aggregation of multiple embedding $\X^{(v)}$ is done implicitly. With this setting, one can easily identify the disagreement degree of different views and tune their weights via the weight learning paradigm.

Considering there are two types of variables to determine in Eq.~\eqref{eq:mvmds}: the configuration points $\X$ and the view weight $\alpha^{(v)}$, we adopt an alternative way to iteratively solve the above optimization problem. By doing so, we decompose it into two sub-problems.

\subsubsection{I. Update $\X$ when $\alpha$ is fixed.}
In this situation, Eq.~\eqref{eq:mvmds} is equivalent to the following optimization problem:
\begin{equation}  \label{eq:J=J1J2J3}
\min_{\X}\J_1+\J_2-2\J_3,
\end{equation}
where
\begin{equation} \label{eq:J1J2J3}
\begin{split}
   &\J_1=\sum_{v=1}^M\sum_{i<j}{ {\alpha^{(v)}}^\gamma w_{ij}{\delta_{ij}^{(v)}}^2 }, \\
   &\J_2=\sum_{v=1}^M\sum_{i<j}{ {\alpha^{(v)}}^\gamma w_{ij}d_{ij}^2(\X) }, \\
   &\J_3=\sum_{v=1}^M\sum_{i<j}{ {\alpha^{(v)}}^\gamma w_{ij}\delta_{ij}^{(v)}d_{ij}(\X) }.
\end{split}
\end{equation}
To optimize this sub-problem, we adopt majorization approach.

As can be drawn, the first term $\J_1$ in Eq.~\eqref{eq:J=J1J2J3} is a constant. Thus it can be omitted in the procedure of optimization.

We now come to the second term in Eq.~\eqref{eq:J=J1J2J3}, which calculates a sum of the weighted squared distances on all views. We can derive that
\begin{equation}
\J_2=trace(\X'\V\X),
\end{equation}
where $\V\in\mathbb{R}^{N\times N}$ has elements
\begin{equation}
v_{ij}=
\begin{cases}
   -\sum_{v=1}^M{\alpha^{(v)}}^\gamma w_{ij} & \textit{if}~i\neq j,\\
   \sum_{j=1,j\neq i}^N{\sum_{v=1}^M{\alpha^{(v)}}^\gamma w_{ij}} & \textit{if}~i=j.
\end{cases}
\end{equation}

The last term in Eq.~\eqref{eq:J=J1J2J3} computes a weighted sum of the distances on all views. Assume $\Z$ denotes the configuration points $\X$ in the previous iteration. According to Cauchy-Schwartz inequality
$d_{ij}(\X){d_{ij}(\Z)}\geq\sum_{p=1}^P(x_{ip}-x_{jp})(z_{ip}-z_{jp})$
with equality if $\Z=\X$, we can obtain
\begin{equation}
\begin{split}
\J_3=\sum_{i<j}{\left(\sum_{v=1}^M{\alpha^{(v)}}^\gamma w_{ij}\delta_{ij}^{(v)}\right)d_{ij}(\X)}\geq trace(\X'\B\Z),
\end{split}
\end{equation}
where $\B\in\mathbb{R}^{N\times N}$ has elements
\begin{equation}
\begin{split}
&b_{ij}=
\begin{cases}
   -\frac{\sum_{v=1}^M{\alpha^{(v)}}^\gamma w_{ij}\delta_{ij}^{(v)}}{d_{ij}(\Z)} &\textit{if}~i\neq j~\textit{and}~d_{ij}(\Z)\neq0\\
   0 & \textit{if}~i\neq j~\textit{and}~d_{ij}(\Z)=0 \\
\end{cases}\\
&b_{ii}=-\sum_{j=1,j\neq i}^N{b_{ij}},
\end{split}
\end{equation}

Based on the analysis above, the objective function in Eq.~\eqref{eq:J=J1J2J3} is upper-bounded by
\begin{equation}
\J\leq \J_{\bigtriangleup}=\J_1+trace(\X'\V\X)-2trace(\X'\B\Z).
\end{equation}
The partial derivative of $J_{\bigtriangleup}$ with regard to $\X$ is
\begin{equation} \label{eq:J_to_X}
\frac{\partial \J_{\bigtriangleup}}{\partial \X}=2\V\X-2\B\Z.
\end{equation}
By setting Eq.~\eqref{eq:J_to_X} to zero, we have
\begin{equation} \label{eq:X1}
\X=\V^{+}\B\Z,
\end{equation}
where $\V^{+}$ is the Moore-Penrose inverse of $\V$. In usual cases, there are no missing values in the input distance matrix $\delta$ (\ie,~$\forall i,j, w_{ij}=1$). Consequently, Eq.~\eqref{eq:X1} can be simplified to
\begin{equation} \label{eq:X2}
\X=\frac{1}{N\sum_{v=1}^M{\alpha^{(v)}}^\gamma}\B\Z,
\end{equation}

\subsubsection{II. Update $\alpha^{(v)}$ when $\X$ is fixed.}
For the sake of notation convenience, we re-write the objective function in Eq.~\eqref{eq:mvmds} as
\begin{equation}
\J=\sum_{v=1}^M{\alpha^{(v)}}^\gamma \J^{(v)},
\end{equation}
where $\J^{(v)}=\sum_{i<j}w_{ij}\left(\delta_{ij}^{(v)}-d_{ij}(\X)\right)^2$ denotes the counterpart of the $v$-th view. To get the optimal solution of this sub-problem, we utilize Lagrange Multiplier Method.

Taking the constraint $\sum_{v=1}^M{\alpha^{(v)}}=1$ into consideration, the Lagrange function of $\J$ is
\begin{equation}
L(\J,\lambda)=\sum_{v=1}^M{\alpha^{(v)}}^\gamma \J^{(v)}+\lambda(\sum_{v=1}^M{\alpha^{(v)}}-1),
\end{equation}
whose partial derivative with respect to $\alpha^{(v)}$ is
\begin{equation} \label{eq:JtoAlpha}
\frac{\partial L(\J, \lambda)}{\partial \alpha^{(v)}}={\gamma{\alpha^{(v)}}^{(\gamma-1)} \J^{(v)}}-\lambda. \\
\end{equation}
By setting Eq.~\eqref{eq:JtoAlpha} to zero, we have
\begin{equation}  \label{eq:JtoAlpha_zero}
\alpha^{(v)}=\left(\frac{\lambda}{\gamma \J^{(v)}}\right)^{\frac1{\gamma-1}}.
\end{equation}
After substituting $\alpha^{(v)}$ in Eq.~\eqref{eq:JtoAlpha_zero} into the constraint $\sum_{v=1}^M{\alpha^{(v)}}=1$, the multiplier $\lambda$ is eliminated and the optimal solution of $\alpha^{(v)}$ is obtained finally as
\begin{equation}  \label{eq:alpha}
\alpha^{(v)}=\frac{ \left(\J^{(v)}\right)^{\frac1{1-\gamma}} }{ \sum_{v'=1}^M{\left(\J^{(v')}\right)^{\frac1{1-\gamma}}} }.\\
\end{equation}

Note that Eq.~\eqref{eq:alpha} encounters ``division by zero" when $\gamma=1$. As discussed above, the optimal solution of $\alpha$ in this situation is
\begin{equation}
\begin{split}
&\alpha^{(v)}=
\begin{cases}
   1 & \textit{if}~v=\arg\min_{v'}{\J^{(v')}}\\
   0 & \textit{otherwise}.
\end{cases}\\
\end{split}
\end{equation}
In the limit case $\gamma\rightarrow\infty$, we will get equal weights $\alpha^{(v)}=\frac1M$ for all the views (see also Fig.~\ref{fig:weight}). As a result, only one parameter $\gamma$ is used to control the weight distribution across multiple views in our algorithm. The optimal choice of $\gamma$ depends the complementarity between the input matrices. If rich complementarity exists among views, large $\gamma$ is preferred.

In summary, we present the whole algorithm in Algorithm~\ref{alg:mvmds}.
The convergence of the proposed algorithm is guaranteed. According to Alg.~\ref{alg:mvmds}, when updating $\X$ in the $(t+1)$-th iteration, the objective value of Eq.~\eqref{eq:mvmds} is decreased by the majorization algorithm compared with that of the $t$-th iteration. When updating $\alpha^{(v)}$, a global minimum is expected to generate the optimal solution based on Eq.~\eqref{eq:alpha}. Therefore, by alternatively updating $\X$ and $\alpha^{(v)}$ in an iterative manner, the objective value keeps decreasing. Since Eq.~\eqref{eq:mvmds} is lower-bounded by $0$, convergence can be arrived given enough iterations.

\begin{algorithm}[tb]
\DontPrintSemicolon
\KwData{\\$\delta^{(v)}\in\mathbb{R}^{N\times N},1\leq v\leq M$: the input distance matrix;\\
      $P$: the embedding dimension; \\
      $\gamma$: the weight controller.\\}
\KwResult{\\$\X\in\mathbb{R}^{N\times P}$: the configuration points; \\
      }
\Begin{
    Initialize $\alpha^{(v)}=\frac{1}{M}$;\\
    \Repeat{convergence}
    {
        Update $\X$ using Eq.~\eqref{eq:X1} or Eq.~\eqref{eq:X2}; \\
        Update the weights $\alpha^{(v)}$ using Eq.~\eqref{eq:alpha}; \\
        Update $\Z=\X$; \\
    }
    \KwRet{$\X$}
}
\caption{Multi-View Multidimensional Scaling.\label{alg:mvmds}}
\end{algorithm}

\section{Future Work}
Many questions remain to be investigated further in this new task, for example:

\subsubsection{Missing values.}~In the proposed solution, one can set $w_{ij}=0$ to ignore missing values in the input distance matrices. Some clustering approaches~\cite{wagstaff2004clustering} usually fill missing values by imputation. Since multiple input distance matrices are available here, maybe it is more effective if we can use the existing values in other views to predict the missing values in a certain view. It deserves a careful investigation in the future, since using the interactions among multiple views to predict missing values have not been exploited before to our best knowledge.

\subsubsection{Intrinsic dimension.}~For the sake of data visualization, the embedding dimension of MDS is usually $P=2$ or $P=3$. In a general situation, $P$ should be specified by the users. Some studies~\cite{levina2004maximum} aim at learning an estimator of intrinsic dimension that can sufficiently describe the data distribution. In this paper, different input distance matrices tend to have different intrinsic dimensions. Therefore, the optimal embedding dimension should not only be ``intrinsic", but also ``consensus",~\ie,~shared by multiple views.
It is probably a data-driven problem. Nevertheless, it is still worthy studying.

\subsubsection{Parameter-free.}~Despite the embedding dimension $P$, standard MDS can be deemed as a parameter-free algorithm. When dealing with multiple input distance matrices, the solution given in this paper introduces an additional parameter $\gamma$ to tune their weight distribution. In our experiments, $\gamma$ has to be specified manually or determined by cross validation. It remains an open issue for researchers to design parameter-free algorithms which can fit into various applications.

\subsubsection{Applications.}~MDS has a wide range of applications in many domains~\cite{lin2016heterogeneous,lindenbaum2015multiview}. These applications can be mostly reconsidered in this newly-defined framework. For example, MDS can be used to draw perceptual maps~\cite{bijmolt1999comparison} in marketing, where each brand has thousands of attributes. Traditionally in MDS, these attributes are treated equally. While with the solution given in this framework (\eg,~MVMDS proposed in this paper), the importance of these attributes can be identified simultaneously. In robots localization~\cite{jenkins2004spatio}, distances between items are usually captured by multiple sensors and multiple time periods. It is badly required to do MDS on multiple distance matrices. These practical applications can be further investigated by researchers in specific domains.

\section{Experiments} \label{sec:exp}
MDS usually acts as a fundamental tool for preprocessing~\cite{IDSC} or visualization~\cite{buja2008data}. For a long time, the only principled way to evaluate the effectiveness of MDS-related algorithms is to compare the stress value defined in Eq.~\eqref{eq:mds}. However, it is not applicable in this paper, owing to the use of multiple groundtruth distances. In this section, we first demonstrate the effectiveness of MVMDS using a synthetic example where multiple views are imitated from a unique groundtruth. Thus, it becomes feasible to compare the stress value using Eq.~\eqref{eq:mds}.
Then following~\cite{MVDR1}, we assess the discriminative power of the embedding $\X$ obtain by MVMDS on three image datasets in the applications of retrieval and clustering.

Since the weight controller $\gamma$ needs to be determined empirically, we conduct an exhaustive search in the interval $(1,10]$ with step size $0.5$ to find its optimal value.

\subsection{Synthetic Example} \label{sec:synthetic}
We consider a synthetic example where $4$ participants are asked to estimate the distances between six cities in the USA, including Los Angeles (LA), San Francisco (SFO), Houston (HOU), Washington D.C. (WC), Chicago (CHI) and New York (NY). Table~\ref{table:city_distance} gives the true pairwise distances among them. Due to the differences in skill and character, different participants generate different estimating results, serving as multiple views. Specially, more professional and careful participants are more likely to attain faithful results.
\begin{table}[tb]
\centering
\begin{tabular}{lcccccc}
\toprule
             & LA  & SFO & CHI & HOU  & NY & WC   \\
\midrule
LA  & 0            & -           & -  & -     & -     & -         \\
SFO  & 380          & 0             & -   & -     & -	     & -          \\
CHI & 2034            & 2148           & 0   & -     & -     & -       \\
HOU  & 1566            & 1945           & 1085   & 0    & -  & -        \\
NY  & 2824            & 2946           & 821   & 1653     & 0     & -       \\
WC  & 2689            & 2840          & 715   & 1414     & 237     & 0          \\
\bottomrule
\end{tabular}
\caption{The pairwise distances among six cities in the USA.}
\label{table:city_distance}
\end{table}

The procedure of generating multiple view input is as follow. To generate the $v$-th view, we first randomly select $K$ pairs of city distances $\delta_{ij}$. Then for each $\delta_{ij}$, Gaussian noise with mean $\delta_{ij}$ and standard derivation $\sigma\cdot\delta_{ij}$ is added. Finally, $4$ views are generated and Table~\ref{table_K} lists the values of $K$ and $\sigma$. As we can see, View $1$ imitates the most proficient and careful participant, since it has the fewest perturbed distance pairs and the smallest derivation. By contrast, View $4$ is the most unskilled and careless participant with lots of mistakes during estimating the distances.
\begin{table}[tb]
\centering
\begin{tabular}{p{2.5cm}p{0.5cm}<{\centering}p{0.5cm}<{\centering}p{3cm}<{\centering}}
\toprule
            & $K$ & $\sigma$  & Stress value ($\times10^5$) \\
\midrule
View 1         & 4     & 0.3     & 2.18      \\
View 2         & 4     & 0.7     & 4.40     \\
View 3         & 8     & 0.3     & 7.50     \\
View 4         & 8     & 0.7     & 74.11     \\
LC\_MDS         & -      & -     & 6.15 \\
MVMDS ($\gamma$=1.5)    & - & -  & 1.61    \\
MVMDS ($\gamma$=5)    & - & -  & \textbf{1.35}    \\
MVMDS ($\gamma$=10)    & - & -  & 1.36    \\
\bottomrule
\end{tabular}
\caption{The parameter setup to generate multi-view input and the comparison of stress.}
\label{table_K}
\end{table}

\begin{figure*}[tb]
\centering
\subfigure[View 1]
{
\begin{minipage}[tb]{0.23\textwidth}
\includegraphics[width = 1\textwidth]{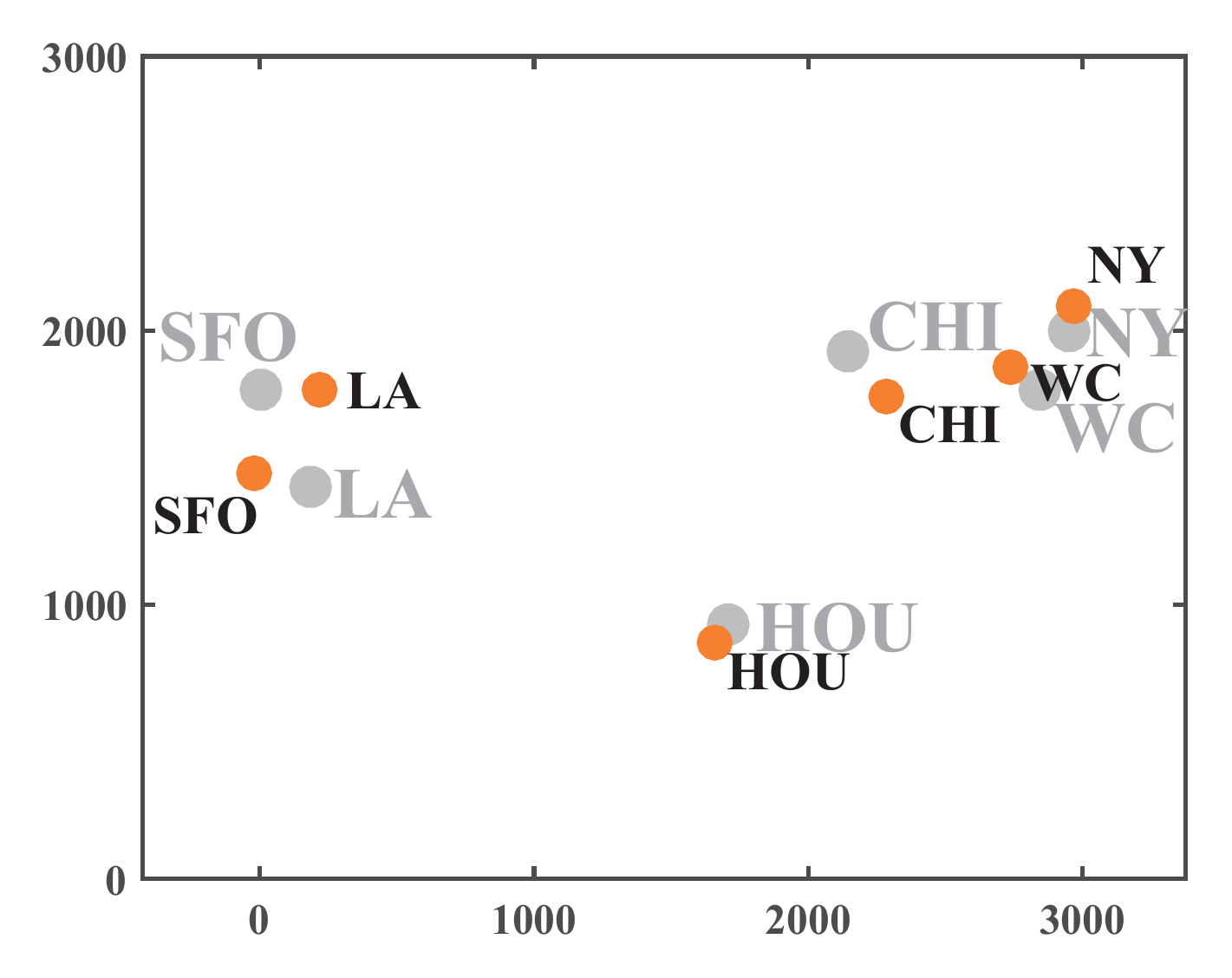}
\end{minipage}
\label{fig:v1}
}
\hspace{-1ex}
\subfigure[View 2]
{
\begin{minipage}[tb]{0.23\textwidth}
\includegraphics[width = 1\textwidth]{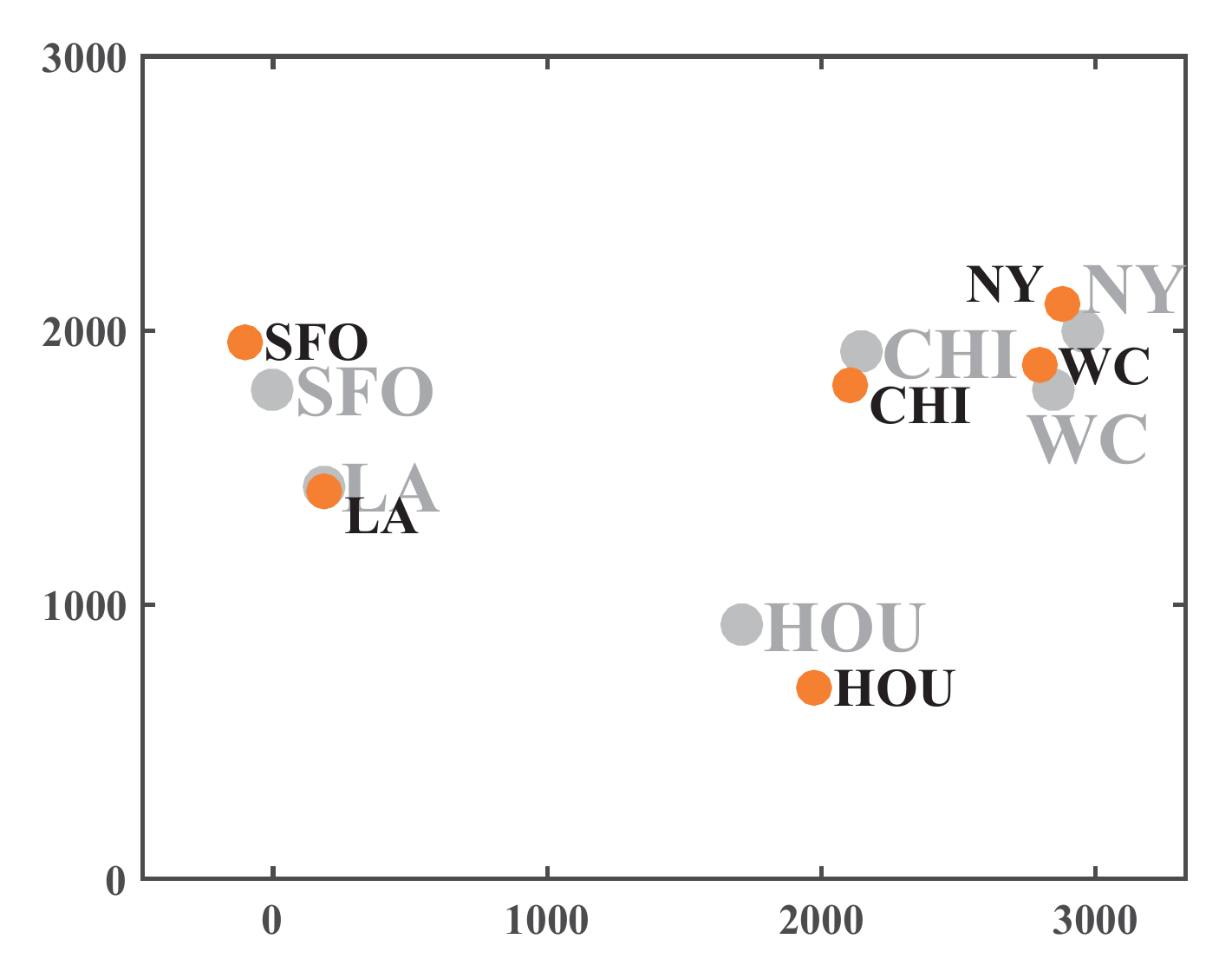}
\end{minipage}
\label{fig:v2}
}
\hspace{-1ex}
\subfigure[View 3]
{
\begin{minipage}[tb]{0.23\textwidth}
\includegraphics[width = 1\textwidth]{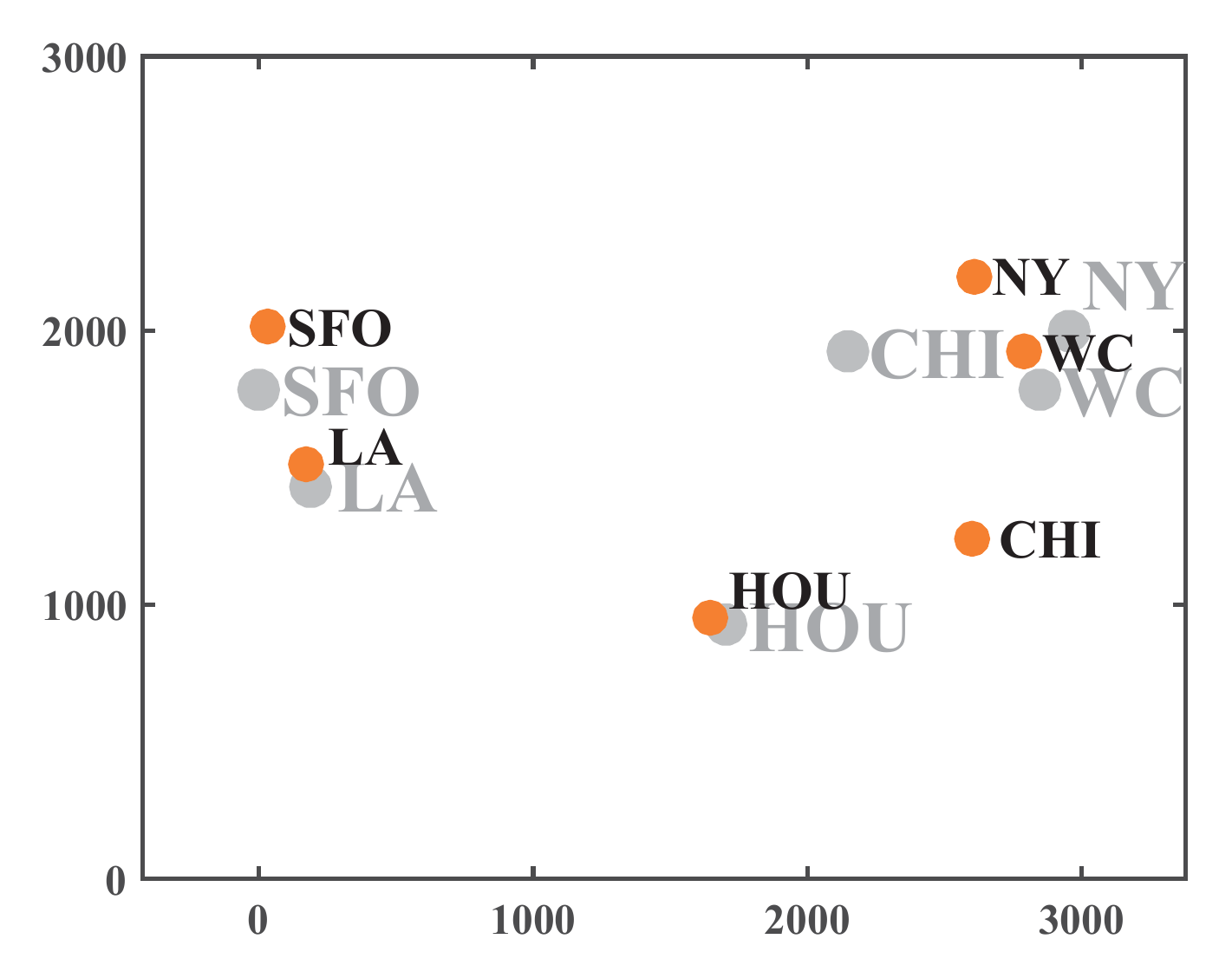}
\end{minipage}
\label{fig:v3}
}
\hspace{-1ex}
\subfigure[View 4]
{
\begin{minipage}[tb]{0.23\textwidth}
\includegraphics[width = 1\textwidth]{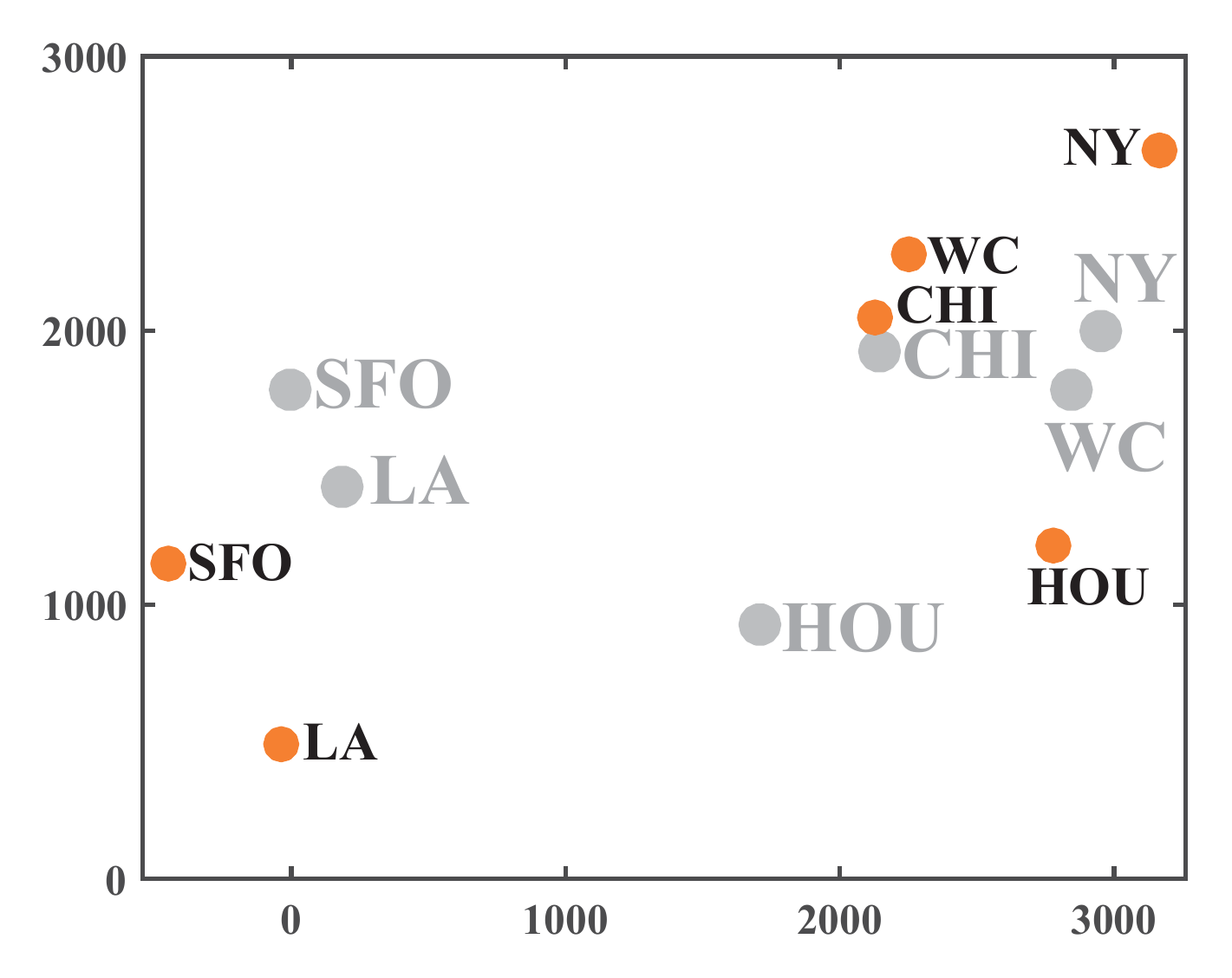}
\end{minipage}
\label{fig:v4}
}

\subfigure[LC\_MDS]
{
\begin{minipage}[tb]{0.23\textwidth}
\includegraphics[width = 1\textwidth]{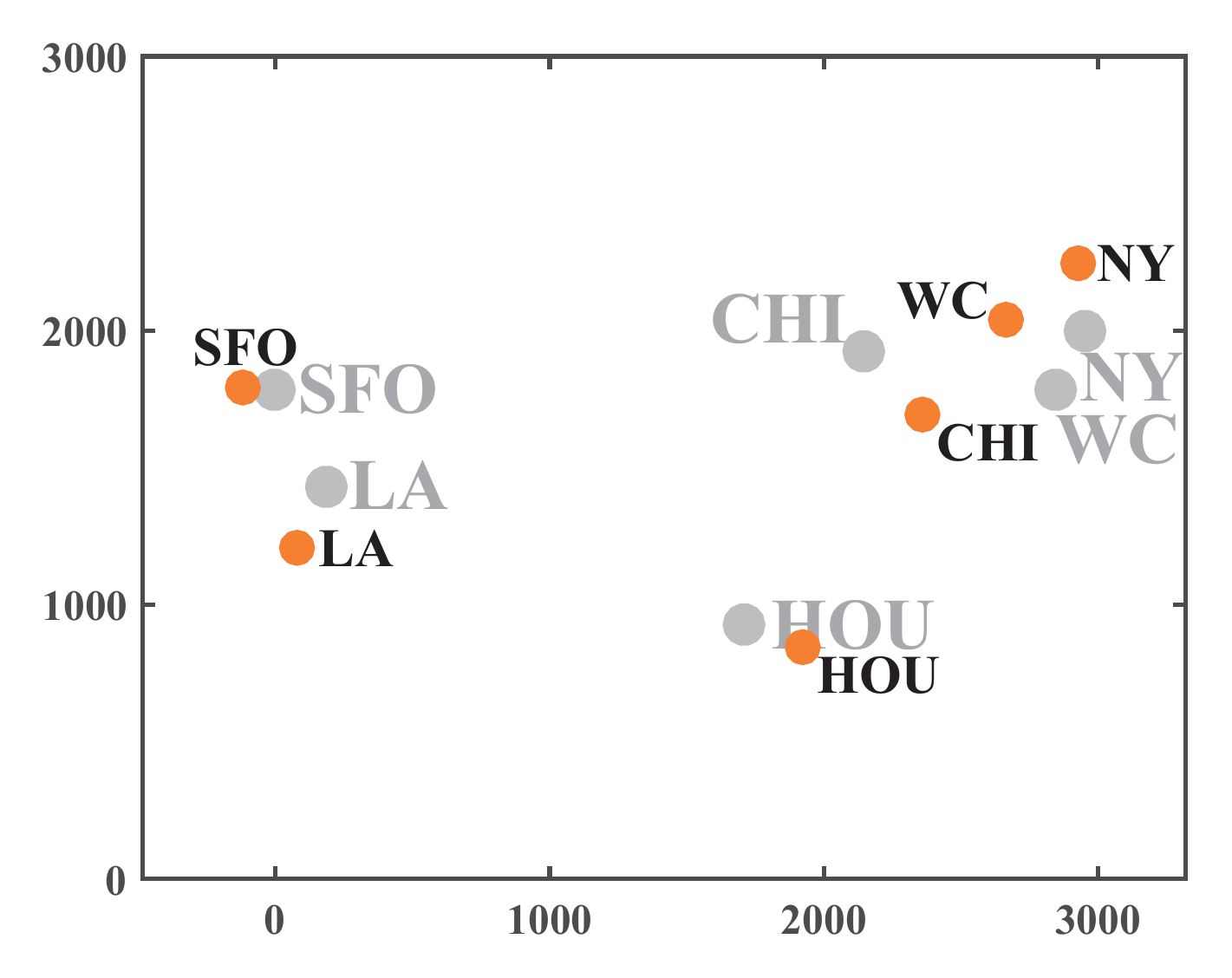}
\end{minipage}
\label{fig:CMDS}
}
\hspace{-1ex}
\subfigure[MVMDS ($\gamma$=1.5)]
{
\begin{minipage}[tb]{0.23\textwidth}
\includegraphics[width = 1\textwidth]{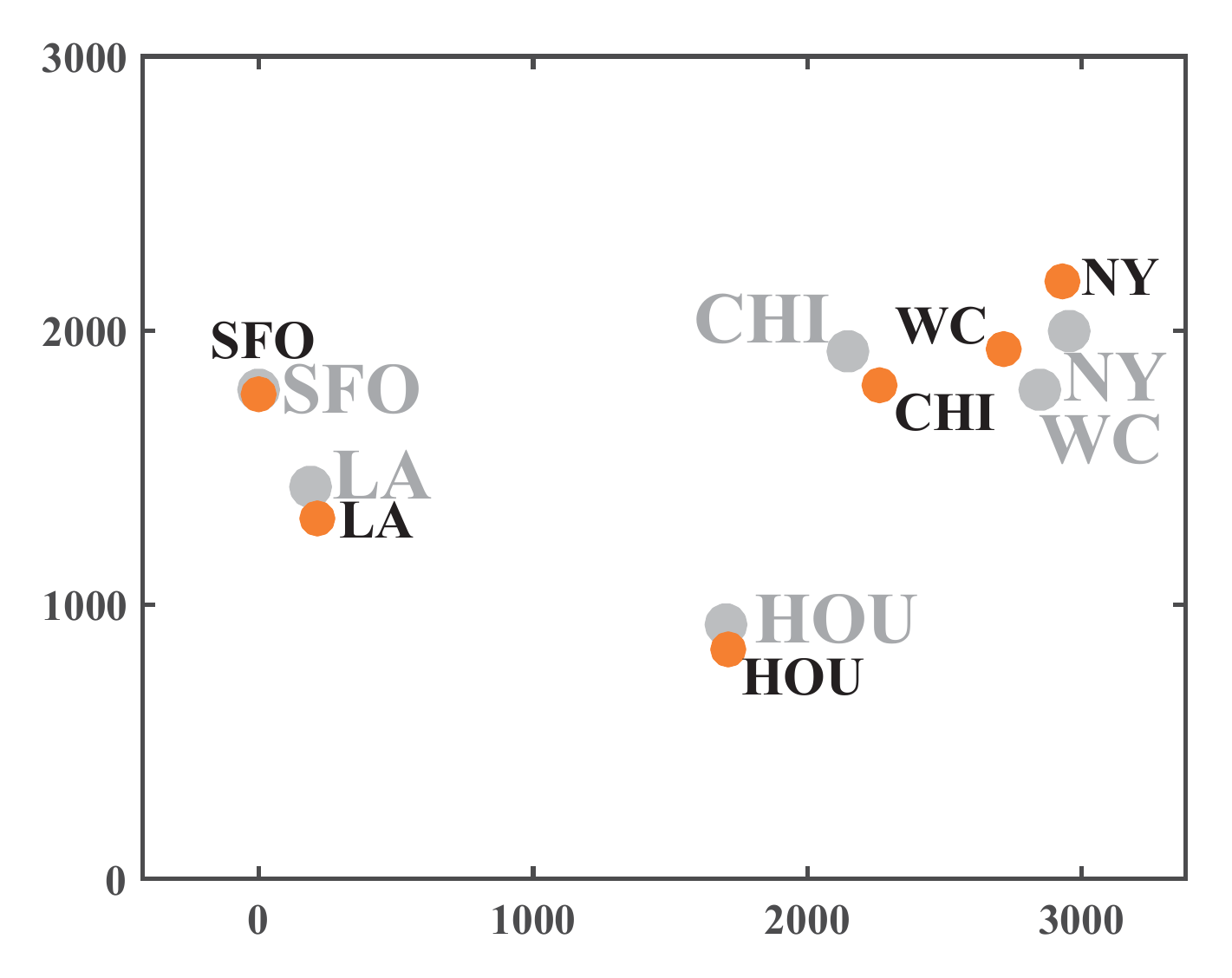}
\end{minipage}
\label{fig:r1}
}
\hspace{-1ex}
\subfigure[MVMDS ($\gamma$=5)]
{
\begin{minipage}[tb]{0.23\textwidth}
\includegraphics[width = 1\textwidth]{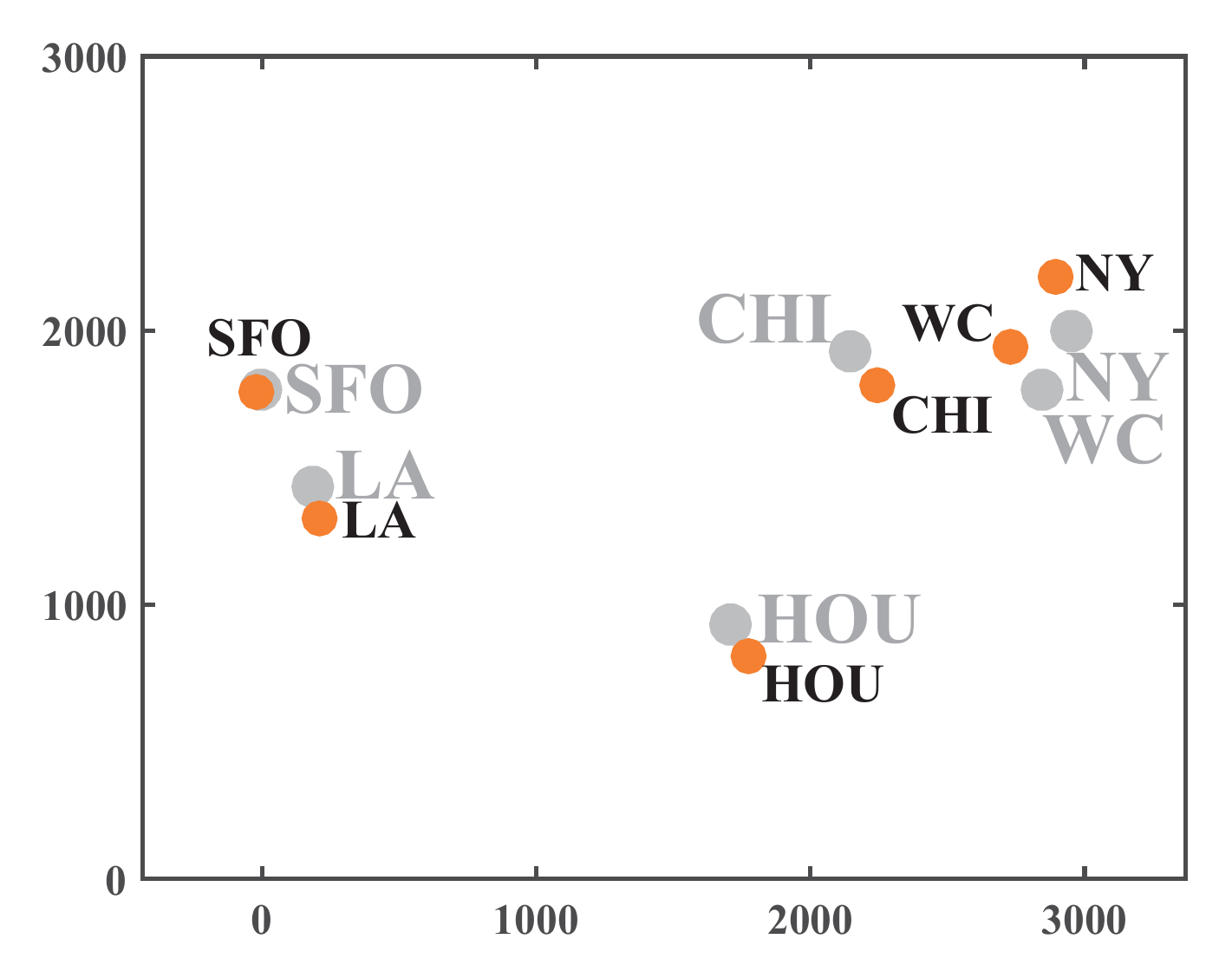}
\end{minipage}
\label{fig:r2}
}
\hspace{-1ex}
\subfigure[MVMDS ($\gamma$=10)]
{
\begin{minipage}[tb]{0.23\textwidth}
\includegraphics[width = 1\textwidth]{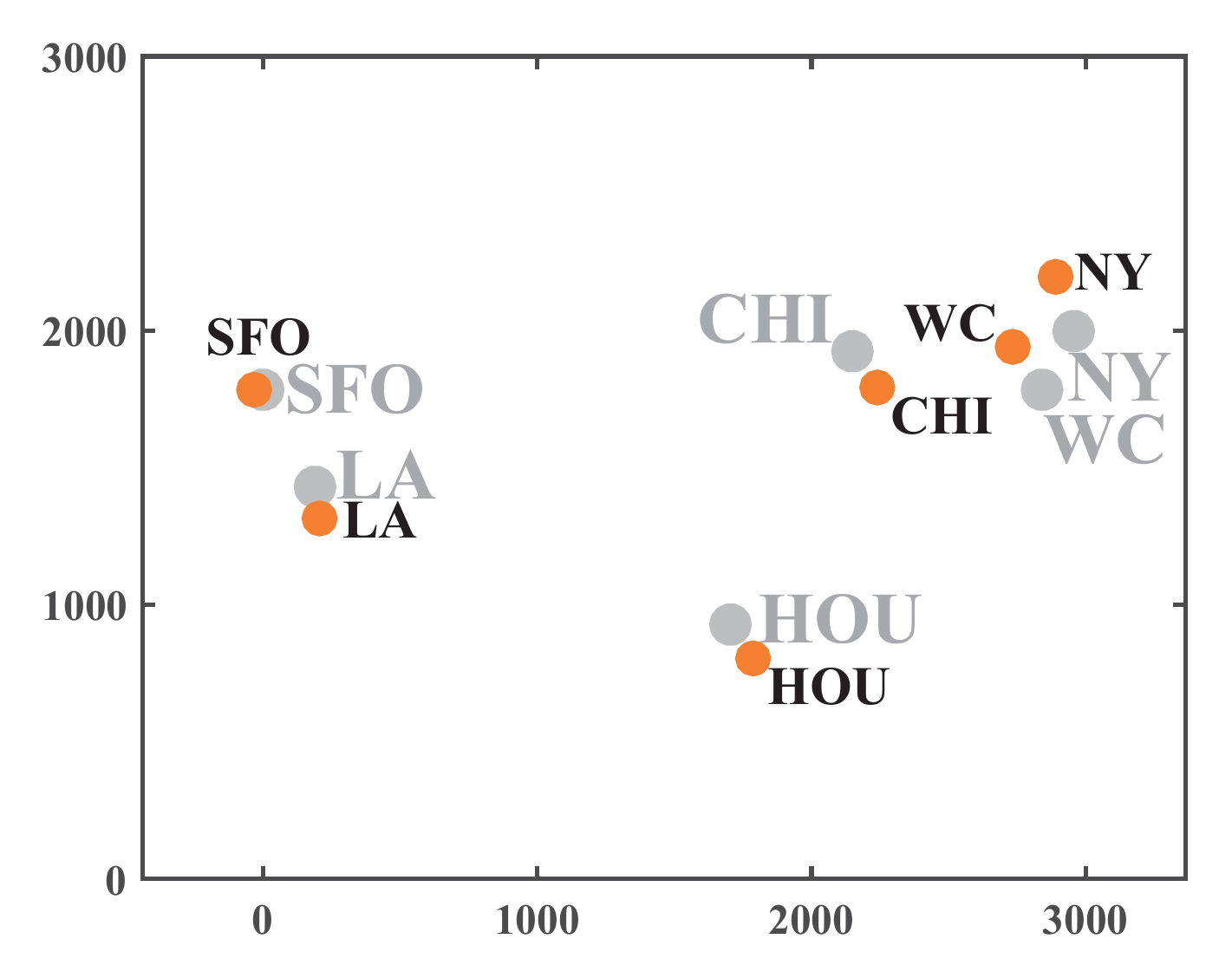}
\end{minipage}
\label{fig:r3}
}
\caption{The illustration of relative positions of the six cities. The gray points are obtained by applying MDS to the true distance matrix shown in Table~\ref{table:city_distance}, serving as groundtruth. The orange points are obtained by applying MDS or MVMDS to simulated views.}
\label{fig:illustration}
\end{figure*}

Fig.~\ref{fig:v1} to Fig.~\ref{fig:v4} give the relative positions of the six cities, marked in orange points, in $P=2$ dimensional space by applying MDS to each view. The result of a linear combination of the $4$ views with equal weights, denoted as LC\_MDS, is presented in Fig.~\ref{fig:CMDS}, and the results of the proposed MVMDS with different $\gamma$ are presented in Fig.~\ref{fig:r1} to Fig.~\ref{fig:r3}. We apply MDS to the distance matrix given in Table~\ref{table:city_distance} to produce the groundtruth, marked in gray color in Fig.~\ref{fig:illustration}. As can be drawn from the figure, MVMDS can yield near perfect results.

\begin{figure}[tb]
\centering
\includegraphics[width=0.92\linewidth]{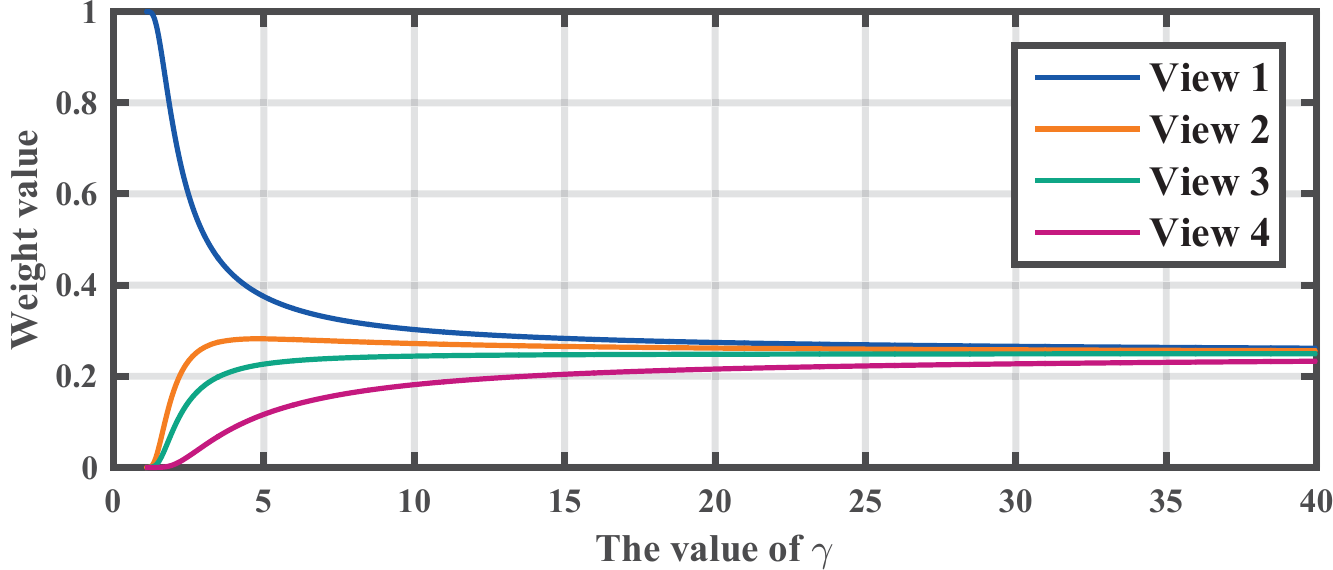}
\caption{The learned weight of different views.}
\label{fig:weight}
\end{figure}

Moreover, since the true distance matrix is accessible in Table~\ref{table:city_distance}, we can directly compute the stress values of different methods using Eq.~\eqref{eq:mds}. The quantitative comparison of stress values is listed in Table~\ref{table_K}. As we can see, the stress of MVMDS is not only lower than each single view but also lower than LC\_MDS.
The reason behind the superiority of MVMDS is the weight learning mechanism imposed on multiple views. To support our claim more clearly, we plot the learned weight $\alpha$ as a function of $\gamma$ in Fig.~\ref{fig:weight}. It suggests that in all the cases, MVMDS can give prominence to View $1$ which is the most reliable participant. When $\gamma<1.5$, the influence of View $4$ is eliminated totally. When $\gamma>35$, we will get equal weights for all the views.

\begin{figure}[tb]
\centering
\includegraphics[width=0.9\linewidth]{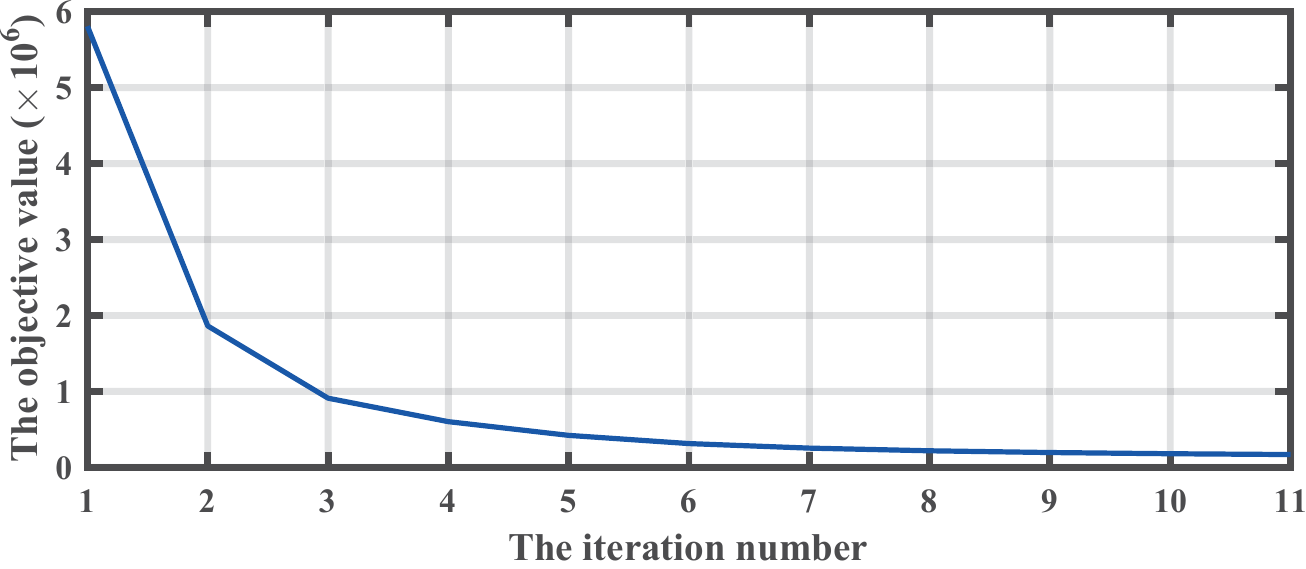}
\caption{The curve of convergence.}
\label{fig:convergence}
\end{figure}
Fig.~\ref{fig:convergence} presents the curve of convergence of MVMDS, which testifies its convergence property experimentally. It is also observed that MVMDS converges quickly within less than $10$ iterations.

\subsection{Image Retrieval} \label{sec:retrieval}
\begin{table*}[tb]
\centering
\small
\begin{tabular}{p{2cm}*{12}{p{0.85cm}<{\centering}}}
\toprule
\multirow{2}{*}{Methods} & \multicolumn{4}{c}{MSRC-v1 dataset} & \multicolumn{4}{c}{Caltech101-7 dataset} & \multicolumn{4}{c}{Caltech101-20 dataset} \\
\cmidrule(lr){2-5} \cmidrule(lr){6-9} \cmidrule(l){10-13}
                         &  NN    & FT     &  ST  & DCG  &  NN    & FT     &  ST  & DCG        &  NN    & FT     &  ST  & DCG   \\
\midrule
SIFT                     & 0.719 & 0.465 & 0.673 & 0.783        & 0.716 & 0.432 & 0.633 & 0.794     & 0.261 & 0.169 & 0.283 & 0.577  \\
HOG                      & 0.728 & 0.481 & 0.681 & 0.790        & 0.788 & 0.515 & 0.700 & 0.832     & 0.417 & 0.261 & 0.384 & 0.636  \\
LBP                      & 0.733 & 0.477 & 0.681 & 0.793        & 0.671 & 0.446 & 0.604 & 0.779     & 0.354 & 0.222 & 0.344 & 0.609  \\
HSV                      & 0.518 & 0.307 & 0.489 & 0.675        & 0.446 & 0.283 & 0.479 & 0.683     & 0.286 & 0.202 & 0.305 & 0.585  \\
GIST                     & 0.742 & 0.460 & 0.663 & 0.786        & 0.721 & 0.496 & 0.677 & 0.810     & 0.425 & 0.262 & 0.373 & 0.636  \\
LC\_MDS                   & 0.796 & 0.501 & 0.705 & 0.816        & 0.711 & 0.460 & 0.651 & 0.797     & 0.304 & 0.216 & 0.314 & 0.594  \\
MVMDS                    & \textbf{0.806} & \textbf{0.530} & \textbf{0.728} & \textbf{0.827}       & \textbf{0.805} & \textbf{0.554} & \textbf{0.734} & \textbf{0.847}     & \textbf{0.429} & \textbf{0.267} & \textbf{0.392} & \textbf{0.641}  \\
\bottomrule
\end{tabular}
\caption{The retrieval performance comparison on MSRC-v1 dataset, Caltech101-7 dataset and Caltech101-20 dataset.}
\label{table_retrieval}
\end{table*}

\begin{table*}[tb]
\small
\centering
\begin{tabular}{lccccccccc}
\toprule
\multirow{2}{*}{Methods} & \multicolumn{3}{c}{MSRC-v1 dataset} & \multicolumn{3}{c}{Caltech101-7 dataset} & \multicolumn{3}{c}{Caltech101-20 dataset} \\
\cmidrule(lr){2-4} \cmidrule(lr){5-7} \cmidrule(l){8-10}
                &  ACC    & NMI    &  Purity     &  ACC    & NMI     &  Purity &  ACC    & NMI     &  Purity   \\
\midrule
SIFT            &61.7$\pm$3.04 &52.8$\pm$2.87 &63.8$\pm$2.85        &55.9$\pm$2.29	&45.6$\pm$2.75 &62.0$\pm$2.19		&24.4$\pm$1.67 &25.6$\pm$1.93 &29.4$\pm$1.87     \\
HOG             &63.6$\pm$1.72 &57.0$\pm$2.08 &65.6$\pm$1.62        &63.5$\pm$1.40	&52.5$\pm$1.82 &68.9$\pm$1.27		&37.4$\pm$1.56 &36.4$\pm$1.58 &41.1$\pm$1.58    \\
LBP             &64.3$\pm$1.37 &57.2$\pm$1.12 &66.7$\pm$1.23        &56.5$\pm$1.24	&43.3$\pm$1.23 &63.4$\pm$1.03		&33.0$\pm$1.20 &33.1$\pm$1.11 &38.9$\pm$1.23  \\
HSV             &42.3$\pm$1.50 &30.7$\pm$1.71 &44.6$\pm$1.46        &32.8$\pm$0.93	&16.8$\pm$0.65 &42.6$\pm$0.69		&26.2$\pm$0.70 &26.5$\pm$0.49 &30.7$\pm$0.63  \\
GIST            &60.2$\pm$1.74 &52.8$\pm$2.02 &63.1$\pm$1.69        &59.2$\pm$1.28	&48.2$\pm$1.09 &63.4$\pm$0.98		&37.6$\pm$1.12 &36.4$\pm$0.86 &41.0$\pm$1.10   \\
LC\_MDS         &70.3$\pm$2.82 &61.9$\pm$3.61 &72.4$\pm$2.81        &56.7$\pm$2.68	&44.8$\pm$2.98 &63.2$\pm$2.44		&27.9$\pm$1.80 &28.4$\pm$1.84 &32.9$\pm$1.72  \\
MVMDS           &\textbf{71.9$\pm$1.82} &\textbf{64.9$\pm$2.68} &\textbf{73.8$\pm$1.89}  &\textbf{71.5$\pm$1.64}&\textbf{63.0$\pm$1.97}&\textbf{76.1$\pm$1.32}		&\textbf{38.5$\pm$1.86} &\textbf{37.6$\pm$1.59} &\textbf{42.3$\pm$1.74}  \\
\bottomrule
\end{tabular}
\caption{The clustering performance comparison (\%) on MSRC-v1 dataset, Caltech101-7 dataset and Caltech101-20 dataset.}
\label{table_clustering}
\end{table*}
Two image benchmark datasets,~\ie,~Microsoft Research Cambridge Volume 1 (MSRC-v1)~\cite{MSRC}, Caltech-101 dataset~\cite{Caltech101}, are selected for performance comparisons. The details of those datasets are listed below:
\begin{enumerate}
  \item MSRC-v1: it is a scene image dataset composed of $240$ images and $9$ categories. Following~\cite{ijcv/LeeG09}, $7$ categories (tree, building, airplane, cow, face, car, bicycle) are used with $30$ images per category.
  \item Caltech-101: it consists of $101$ object categories, with $31$ to $800$ images per category. Following~\cite{AP_ICCV}, we select $7$ classes and $20$ classes forming Caltech101-7 and Caltech101-20 respectively.
\end{enumerate}

We extract $5$ visual features to obtain the multi-view representations for each image,~\ie,~Scale Invariant Feature Transform (SIFT)~\cite{SIFT} with dimension $128$, Histogram of Oriented Gradients (HOG)~\cite{HOG} with dimension $775$, Local Binary Patterns (LBP)~\cite{LBP} with dimension $1450$, HSV color histogram with dimension $1000$, GIST~\cite{GIST} with dimension $512$. All the visual features are $L_2$ normalized, then Euclidean distance is used to measure the dissimilarity between images.
To get a comprehensive quantitative evaluation, we adopt four widely-used metrics in information retrieval,~\ie,~Nearest Neighbor (NN), First Tier (FT), Second Tier (ST) and Discounted Cumulative Gain (DCG). All the metrics range from $0$ to $1$ and larger values indicate better performances. Please refer to~\cite{PSB} for their detailed definitions.

We compare the proposed MVMDS against $6$ methods, including $5$ single view counterparts and LC\_MDS. All the comparisons are done by using MDS or the proposed MVMDS to project images into $P=10$ dimensional space. Table~\ref{table_retrieval} presents the experimental results on all the datasets. The table shows that our proposed MVMDS achieves the best performances consistently in all the evaluation metrics. One can also find that using a linear combination of all the views is not always useful. For example, the performances of LC\_MDS are much lower than those of HOG on Caltech101-7 dataset and Caltech101-20 dataset. Our interpretation is that the baseline performances of most views (\eg,~SIFT, LBP and HSV) are poor, and they will deprive the discriminative power of informative views (\eg,~HOG and GIST) by simply stacking them with equal weights. By contrast, the proposed MVMDS benefits from the weight learning paradigm, thus decreasing the weights of less information views and suppressing their negative effects to a certain extent.

\subsection{Image Clustering} \label{sec:clustering}
In this section, we evaluate the performances of MVMDS in clustering task to obtain a more thorough analysis.
We also extract $5$ visual features and project all images into $P=10$ dimensional space. Then K-means is applied to divide the images into clusters. The desired number of clusters is set to be equal to the natural number of categories in each dataset. For performance evaluation, we adopt three widely-used evaluation metrics, that is, Clustering Accuracy (ACC), Normalized Mutual Information (NMI) and Purity.

The comparison is presented in Table~\ref{table_clustering}. Consistent to the experimental results above, MVMDS outperforms all the compared methods by a large margin. Especially on Caltech101-7 dataset, MVMDS outperforms the best-performing single view (HOG) by $7.97\%$ in ACC, $10.51\%$ in NMI, $7.18\%$ in Purity and LC\_MDS by $14.76\%$ in ACC, $18.16\%$ in NMI, $12.87\%$ in Purity respectively.

We also compare with other multi-view learning algorithms, though they are not MDS-based. For example, the performance of MVMDS is better than Robust Multi-view K-means Clustering (RMKMC)~\cite{cai2013multi}, which reports ACC $67.9$, NMI $68.9$ and Purity $75.9$. The performance gain is especially valuable when considering that the feature dimension used by MVMDS is only $P=10$, significantly shorter than $2346$ dimensional feature used in RMKMC.

\section{Conclusion} \label{sec:con}
In this paper, we focus on a new problem, that is, performing Multidimensional Scaling (MDS) on multi-view data. To address this issue, we propose a new algorithm called Multi-View Multidimensional Scaling (MVMDS), which is optimized in an iterative manner with guaranteed convergence. The proposed method can do discriminative view selection adaptively, thus the contributions of informative views are amplified. As introduced above, there are many interesting problems and applications for following researchers to think deeply in the future.


\bibliographystyle{aaai}
\bibliography{MVMDS}

\begin{thebibliography}{}

\bibitem[\protect\citeauthoryear{Amid and Ukkonen}{2015}]{amid2015multiview}
Amid, E., and Ukkonen, A.
\newblock 2015.
\newblock Multiview triplet embedding: Learning attributes in multiple maps.
\newblock In {\em ICML},  1472--1480.

\bibitem[\protect\citeauthoryear{Bijmolt and
  Wedel}{1999}]{bijmolt1999comparison}
Bijmolt, T.~H., and Wedel, M.
\newblock 1999.
\newblock A comparison of multidimensional scaling methods for perceptual
  mapping.
\newblock {\em Journal of Marketing Research}  277--285.

\bibitem[\protect\citeauthoryear{Borg and Groenen}{2005}]{borg2005modern}
Borg, I., and Groenen, P.~J.
\newblock 2005.
\newblock {\em Modern multidimensional scaling: Theory and applications}.
\newblock Springer Science \& Business Media.

\bibitem[\protect\citeauthoryear{Bronstein \bgroup et al\mbox.\egroup
  }{2008}]{bronstein2008analysis}
Bronstein, A.~M.; Bronstein, M.~M.; Bruckstein, A.~M.; and Kimmel, R.
\newblock 2008.
\newblock Analysis of two-dimensional non-rigid shapes.
\newblock {\em IJCV} 78(1):67--88.

\bibitem[\protect\citeauthoryear{Buja \bgroup et al\mbox.\egroup
  }{2008}]{buja2008data}
Buja, A.; Swayne, D.~F.; Littman, M.~L.; Dean, N.; Hofmann, H.; and Chen, L.
\newblock 2008.
\newblock Data visualization with multidimensional scaling.
\newblock {\em Journal of Computational and Graphical Statistics}
  17(2):444--472.

\bibitem[\protect\citeauthoryear{Cai, Nie, and Huang}{2013}]{cai2013multi}
Cai, X.; Nie, F.; and Huang, H.
\newblock 2013.
\newblock Multi-view k-means clustering on big data.
\newblock In {\em IJCAI}.

\bibitem[\protect\citeauthoryear{Dalal and Triggs}{2005}]{HOG}
Dalal, N., and Triggs, B.
\newblock 2005.
\newblock Histograms of oriented gradients for human detection.
\newblock In {\em CVPR},  886--893.

\bibitem[\protect\citeauthoryear{Dueck and Frey}{2007}]{AP_ICCV}
Dueck, D., and Frey, B.~J.
\newblock 2007.
\newblock Non-metric affinity propagation for unsupervised image
  categorization.
\newblock In {\em ICCV},  1--8.

\bibitem[\protect\citeauthoryear{Elad and Kimmel}{2003}]{elad2003bending}
Elad, A., and Kimmel, R.
\newblock 2003.
\newblock On bending invariant signatures for surfaces.
\newblock {\em TPAMI} 25(10):1285--1295.

\bibitem[\protect\citeauthoryear{Fei-Fei, Fergus, and
  Perona}{2007}]{Caltech101}
Fei-Fei, L.; Fergus, R.; and Perona, P.
\newblock 2007.
\newblock Learning generative visual models from few training examples: An
  incremental bayesian approach tested on 101 object categories.
\newblock {\em CVIU} 106(1):59--70.

\bibitem[\protect\citeauthoryear{Foster, Kakade, and
  Zhang}{2008}]{foster2008multi}
Foster, D.~P.; Kakade, S.~M.; and Zhang, T.
\newblock 2008.
\newblock Multi-view dimensionality reduction via canonical correlation
  analysis.
\newblock In {\em Technical Report}.

\bibitem[\protect\citeauthoryear{France and Carroll}{2011}]{france2011two}
France, S.~L., and Carroll, J.~D.
\newblock 2011.
\newblock Two-way multidimensional scaling: A review.
\newblock {\em IEEE Trans. on Systems, Man, and Cybernetics, Part C:
  Applications and Reviews} 41(5):644--661.

\bibitem[\protect\citeauthoryear{Han \bgroup et al\mbox.\egroup }{2012}]{MVDR1}
Han, Y.; Wu, F.; Tao, D.; Shao, J.; Zhuang, Y.; and Jiang, J.
\newblock 2012.
\newblock Sparse unsupervised dimensionality reduction for multiple view data.
\newblock {\em IEEE Trans. Circuits Syst. Video Techn} 22(10):1485--1496.

\bibitem[\protect\citeauthoryear{Jenkins and
  Matari{\'c}}{2004}]{jenkins2004spatio}
Jenkins, O.~C., and Matari{\'c}, M.~J.
\newblock 2004.
\newblock A spatio-temporal extension to isomap nonlinear dimension reduction.
\newblock In {\em ICML}, ~56.

\bibitem[\protect\citeauthoryear{Lee and Grauman}{2009}]{ijcv/LeeG09}
Lee, Y.~J., and Grauman, K.
\newblock 2009.
\newblock Foreground focus: Unsupervised learning from partially matching
  images.
\newblock {\em IJCV} 85(2):143--166.

\bibitem[\protect\citeauthoryear{Levina and Bickel}{2004}]{levina2004maximum}
Levina, E., and Bickel, P.~J.
\newblock 2004.
\newblock Maximum likelihood estimation of intrinsic dimension.
\newblock In {\em NIPS},  777--784.

\bibitem[\protect\citeauthoryear{Lin \bgroup et al\mbox.\egroup
  }{2016}]{lin2016heterogeneous}
Lin, G.; Fan, G.; Kang, X.; Zhang, E.; and Yu, L.
\newblock 2016.
\newblock Heterogeneous feature structure fusion for classification.
\newblock {\em Pattern Recognition} 53:1--11.

\bibitem[\protect\citeauthoryear{Lindenbaum \bgroup et al\mbox.\egroup
  }{2015}]{lindenbaum2015multiview}
Lindenbaum, O.; Yeredor, A.; Salhov, M.; and Averbuch, A.
\newblock 2015.
\newblock Multiview diffusion maps.
\newblock {\em arXiv preprint arXiv:1508.05550}.

\bibitem[\protect\citeauthoryear{Ling and Jacobs}{2007}]{IDSC}
Ling, H., and Jacobs, D.~W.
\newblock 2007.
\newblock Shape classification using the inner-distance.
\newblock {\em TPAMI} 29(2):286--299.

\bibitem[\protect\citeauthoryear{Lowe}{2004}]{SIFT}
Lowe, D.~G.
\newblock 2004.
\newblock Distinctive image features from scale-invariant keypoints.
\newblock {\em IJCV} 60(2):91--110.

\bibitem[\protect\citeauthoryear{Ojala, Pietik{\"{a}}inen, and
  M{\"{a}}enp{\"{a}}{\"{a}}}{2002}]{LBP}
Ojala, T.; Pietik{\"{a}}inen, M.; and M{\"{a}}enp{\"{a}}{\"{a}}, T.
\newblock 2002.
\newblock Multiresolution gray-scale and rotation invariant texture
  classification with local binary patterns.
\newblock {\em TPAMI} 24(7):971--987.

\bibitem[\protect\citeauthoryear{Oliva and Torralba}{2001}]{GIST}
Oliva, A., and Torralba, A.
\newblock 2001.
\newblock Modeling the shape of the scene: {A} holistic representation of the
  spatial envelope.
\newblock {\em IJCV} 42(3):145--175.

\bibitem[\protect\citeauthoryear{Shilane \bgroup et al\mbox.\egroup
  }{2004}]{PSB}
Shilane, P.; Min, P.; Kazhdan, M.~M.; and Funkhouser, T.~A.
\newblock 2004.
\newblock The princeton shape benchmark.
\newblock In {\em SMI}.

\bibitem[\protect\citeauthoryear{Sun}{2013}]{sun2013survey}
Sun, S.
\newblock 2013.
\newblock A survey of multi-view machine learning.
\newblock {\em Neural Computing and Applications} 23(7-8):2031--2038.

\bibitem[\protect\citeauthoryear{Taguchi and
  Oono}{2005}]{taguchi2005relational}
Taguchi, Y.-h., and Oono, Y.
\newblock 2005.
\newblock Relational patterns of gene expression via non-metric
  multidimensional scaling analysis.
\newblock {\em Bioinformatics} 21(6):730--740.

\bibitem[\protect\citeauthoryear{Torgerson}{1958}]{torgerson1958theory}
Torgerson, W.~S.
\newblock 1958.
\newblock Theory and methods of scaling.

\bibitem[\protect\citeauthoryear{Wagstaff}{2004}]{wagstaff2004clustering}
Wagstaff, K.
\newblock 2004.
\newblock {\em Clustering with missing values: No imputation required}.
\newblock Springer.

\bibitem[\protect\citeauthoryear{Winn and Jojic}{2005}]{MSRC}
Winn, J.~M., and Jojic, N.
\newblock 2005.
\newblock {LOCUS:} learning object classes with unsupervised segmentation.
\newblock In {\em ICCV},  756--763.

\bibitem[\protect\citeauthoryear{Xia \bgroup et al\mbox.\egroup }{2010}]{MVDR2}
Xia, T.; Tao, D.; Mei, T.; and Zhang, Y.
\newblock 2010.
\newblock Multiview spectral embedding.
\newblock {\em IEEE Transactions on Systems, Man, and Cybernetics, Part B}
  40(6):1438--1446.

\bibitem[\protect\citeauthoryear{Xu, Tao, and Xu}{2013}]{xu2013survey}
Xu, C.; Tao, D.; and Xu, C.
\newblock 2013.
\newblock A survey on multi-view learning.
\newblock {\em arXiv preprint arXiv:1304.5634}.

\bibitem[\protect\citeauthoryear{Zhou and Li}{2005}]{co_training2}
Zhou, Z.-H., and Li, M.
\newblock 2005.
\newblock Semi-supervised regression with co-training.
\newblock In {\em IJCAI}, volume~5,  908--913.

\end{thebibliography}

\end{document}